# Attribute Value Weighting in *K*-Modes Clustering

Zengyou He, Xaiofei Xu, Shengchun Deng

Department of Computer Science and Engineering, Harbin Institute of Technology,

92 West Dazhi Street, P.O Box 315, Harbin 150001, P. R. China

zengyouhe@yahoo.com, {xiaofei, dsc}@hit.edu.cn

**Abstract** In this paper, the traditional *k*-modes clustering algorithm is extended by weighting attribute value matches in dissimilarity computation. The use of attribute value weighting technique makes it possible to generate clusters with stronger intra-similarities, and therefore achieve better clustering performance. Experimental results on real life datasets show that these value weighting based *k*-modes algorithms are superior to the standard *k*-modes algorithm with respect to clustering accuracy.

## 1. Introduction

Categorical data clustering is an important research problem in pattern recognition and data mining. The *k*-modes algorithm [1] extends the *k*-means paradigm to cluster categorical data by using (1) a simple matching dissimilarity measure for categorical objects, (2) modes instead of means for clusters, and (3) a frequency-based method to update modes in the *k*-means fashion to minimize the cost function of clustering. The *k*-modes algorithm is widely used in real world applications due to its efficiency in dealing with large categorical database.

In standard *k*-modes algorithm, a simple matching similarity measure is used, in which the distance is either 0 or 1. Such simple matching dissimilarity measure doesn't consider the implicit similarity relationship embedded in categorical values, which will result in a weaker intra-cluster similarity by allocating less similar objects to the cluster. To illustrate this fact, let's consider the following example shown in Fig.1.

*Example 1:* In this artificial example, the dataset is described with 3 categorical attributes *A1*, *A2*, and *A3*, and there are two clusters with their modes. Assuming that we have to allocate a data object $Y$ = [a, p, w] to either cluster 1 or cluster 2. According to the *k*-modes algorithm, we can assign $Y$ to either cluster 1 or cluster 2 since these two clusters have the same mode. However, from the viewpoint of intra-cluster similarity, it is more desirable to allocate $Y$ to cluster 1.

Cluster 1: mode1=[a, p, r]   Cluster 2: mode2=[a, p, r]

| *A1* | *A2* | *A3* |
|------|------|------|
| a    | p    | r    |
| a    | p    | s    |
| a    | p    | t    |

| *A1* | *A2* | *A3* |
|------|------|------|
| a    | q    | r    |
| b    | p    | t    |
| a    | p    | k    |

**Fig. 1** Two clusters with their modes in example 1

The above example shows that the dissimilarity measure used in *k*-modes algorithm treat all attribute values equally, which can not always represent the real semantic distance between a data object and a cluster. To address this problem, this paper extends *k*-modes clustering algorithm by weighting attribute value matches in dissimilarity computation. More precisely, the similarity between two identical categorical values is not always one but a value-dependent weight $\omega$ ranging from 0 to 1. Consequently, the distance between two distinct categorical values is still 1, while the distance between two identical categorical values becomes $1-\omega$. Such modification on distance measure allows the algorithm to recognize clusters with strong intra-similarities, and therefore achieve better clustering performance.

A number of non-binary similarity measures (e.g., [2-6]) have been proposed for categorical values, which can be exploited for providing different weights on attribute value matches in *k*-modes. However, existing methods are not very feasible in our problem since most of them are specially designed for the task of supervised learning [3-4] or manual efforts are needed to specify distance hierarchies [5]. Therefore, this paper presents several frequency-based weighting methods, which result in several variants of attribute value weighting based *k*-modes algorithm.

New *k*-modes algorithms with frequency-based weighting schemas are easy to be implemented with *only* minor modifications on the original *k*-modes algorithm. More importantly, these new algorithms can achieve better clustering accuracies without sacrificing the scalability of original *k*-modes algorithm. Experimental results on real life datasets verify the superiority of our methods.

The rest of this paper is organized as follows: Section 2 reviews the literature from several viewpoints and Section 3 reviews the standard *k*-modes algorithm. Then attribute value weighting based *k*-modes algorithms are presented in Section 4. Finally, Section 5 gives empirical results and Section 6 concludes the paper.

## 2. Literature Review

The *k*-modes algorithm [1] has become a popular technique in solving categorical data clustering problems in different application domains. In this paper, our focus is to study *k*-modes type clustering for categorical data. Hence, we will review only those *k*-modes related papers.

Following the *k*-modes algorithm, many research efforts [7-14] have been conducted to further improve its performance.

Huang and Ng introduced the fuzzy *k*-modes algorithm [7], which assigns membership degrees to data objects in different clusters. The tabu search technique is applied in [8] and genetic algorithm is utilized in [9] to improve *k*-modes algorithm. Alternatively, fuzzy *k*-modes algorithm is extended by representing the clusters of categorical data with fuzzy centroids instead of the hard-type centroids used in the original algorithm [10-11]. However, most of these methods are much slower than the original *k*-modes algorithm in running time.

Since the *k*-modes algorithm is sensitive to the initial conditions, another feasible way for improving its performance is to design effective initialization methods. To that end, an iterative initial-points refinement algorithm for categorical data is presented in [12].

A new *k*-modes algorithm is developed in our previous work [13], which applies a new dissimilarity measure to the *k*-modes clustering. The main idea is to use the relative attribute frequencies of the cluster modes in the similarity measure in the *k*-modes objective function. The method in [13] could be considered as one special case of frequency-based weighting schemas studied in this paper. Ng et al. [14] provide a theoretical understanding of the proposed *k*-modes algorithm in [13].

Measuring similarities between categorical data objects is a difficult task since relations between categorical values cannot be mathematical specified or easily established. Recently, how to define a good distance (dissimilarity) measure between categorical data objects has started drawing more attentions (e.g., [2-6]). However, most existing methods are not very feasible in our problem due to the following reasons. (1) Some techniques [3-4] are specially designed for the task of supervised learning, which are not applicable to cluster analysis. (2) Manual efforts are needed in defining distance hierarchies in [5], which is error-prone and not feasible for attribute that takes on many values. (3) Some recent work [2] has proposed algorithms that can automatically produce distance hierarchies without human interactions, however, these algorithms are very complicated and time-consuming. One exception is the method in [6], which adopts a similarity measure that gives greater weight to uncommon attribute value matches in similarity computations. However, its feasibility in *k*-modes clustering is still unknown.

Attribute weighting is also an important research topic in cluster analysis. In particular, attribute weighting in *k*-means-type clustering has started drawing more attentions (e.g., [15-18]). We remark that our focus in this paper is to weight *attribute values* instead of *attributes*. It seems that it is not an easy task to apply those techniques in [15-18] to attribute value weighting in categorical data clustering.

## 3. The *k*-modes Algorithm

Let $D = \{X_1, X_2, \ldots, X_n\}$ be a set of *n* categorical objects, where each $X_i = [x_{i,1}, x_{i,2}, \ldots, x_{i,m}]$ is described by *m* categorical attributes $A_1, \ldots, A_m$. Each attribute $A_j$ describes a domain of values, denoted by $Dom(A_j) = \{a_j^{(1)}, a_j^{(2)}, \ldots, a_j^{(p_j)}\}$, where $p_j$ is the number of category values of attribute $A_j$.

The *k*-modes algorithm uses the *k*-means paradigm to search a partition of *D* into *k* clusters that minimize the objective function *P* with unknown variables *U* and *Z* as follows:

$$P(U,Z) = \sum_{l=1}^{k} \sum_{i=1}^{n} \sum_{j=1}^{m} u_{i,l} d(x_{i,j}, z_{l,j}) \tag{1}$$

$$\text{Subject to } \sum_{l=1}^{k} u_{i,l} = 1, \quad 1 \leq i \leq n \tag{2}$$

$$u_{i,l} \in \{0,1\}, \quad 1 \leq i \leq n, \quad 1 \leq l \leq k \tag{3}$$

where

- *U* is an $n \times k$ partition matrix, $u_{i,l}$ is a binary variable, and $u_{i,l} = 1$ indicates that object $X_i$ is allocated to cluster $C_l$;
- $Z = \{Z_1, Z_2, \ldots, Z_k\}$ is a set of vectors representing the centers of the *k* clusters, where $Z_l = [z_{l,1}, z_{l,2}, \ldots, z_{l,m}]$ ($1 \leq l \leq k$);

- $d(x_{i,j}, z_{l,j})$ is a distance or dissimilarity measure between object $X_i$ and the center of cluster $C_l$ on attribute $A_j$. In k-modes algorithm, a simple matching distance measure is used. That is, the distance between two distinct categorical values is 1, while the distance between two identical categorical values is 0. More precisely,

$$d(x_{i,j}, z_{l,j}) = \begin{cases} 0 & (x_{i,j} = z_{l,j}) \\ 1 & (x_{i,j} \neq z_{l,j}) \end{cases} \quad (4)$$

The optimization problem in k-modes clustering can be solved by iteratively solving the following two minimization problems:

1. Problem $P_1$: Fix $Z = \hat{Z}$, solve the reduced problem $P(U, \hat{Z})$,

2. Problem $P_2$: Fix $U = \hat{U}$, solve the reduced problem $P(\hat{U}, Z)$.

Problem $P_1$ and $P_2$ are solved according to the two following theorems, respectively.

**Theorem 1**: Let $Z = \hat{Z}$ be fixed, $P(U, \hat{Z})$ is minimized iff

$$u_{i,l} = \begin{cases} 1 & (\sum_{j=1}^{m} d(x_{i,j}, z_{l,j}) \leq \sum_{j=1}^{m} d(x_{i,j}, z_{h,j}), \forall h, 1 \leq h \leq k) \\ 0 & (otherwise) \end{cases}$$

**Theorem 2**: Let $U = \hat{U}$ be fixed, $P(\hat{U}, Z)$ is minimized iff

$$z_{l,j} = a_j^{(r)}$$

where $a_j^{(r)}$ is the mode of attribute values of $A_j$ in cluster $C_l$ that satisfies $f(a_j^{(r)} | C_l) \geq f(a_j^{(t)} | C_l), \forall t, 1 \leq t \leq p_j$. Here $f(a_j^{(r)} | C_l)$ denotes the frequency count of attribute value $a_j^{(r)}$ in cluster $C_l$, i.e., $f(a_j^{(r)} | C_l) = |\{u_{i,l} | x_{i,j} = a_j^{(r)}, u_{i,l} = 1\}|$.

The above idea is formalized in k-modes algorithm as follows.

**Algorithm (The k-modes algorithm)**

1. Randomly choose an initial $Z^{(1)}$. Determine $U^{(1)}$ such that $P(U, Z^{(1)})$ is minimized. Set $t = 1$.

2. Determine $Z^{(t+1)}$ such that $P(U^{(t)}, Z^{(t+1)})$ is minimized. If $P(U^{(t)}, Z^{(t+1)}) = P(U^{(t)}, Z^{(t)})$, then stop; otherwise, go to step 3.

3. Determine $U^{(t+1)}$ such that $P(U^{(t+1)}, Z^{(t+1)})$ is minimized. If $P(U^{(t+1)}, Z^{(t+1)}) = P(U^{(t)}, Z^{(t+1)})$, then stop; otherwise, set $t = t+1$ and go to step 2.

# 4. Attribute Value Weighting Based *k*-modes Algorithms

## 4.1 Problem Formulation

The use of simple matching dissimilarity measure in standard *k*-modes algorithm may cause problems in both *object allocation* and *center selection*, which are summarized as follows.

1. *Object Allocation*: In solving Problem $P_1$ for *k*-modes clustering, data objects are allocated to clusters according to Theorem 1, i.e., each data object is assigned to its nearest cluster. However, the simple matching distance measure is either 0 or 1, which cannot always represent the real semantic distance between a data object and a cluster. As shown in the Example of Section 1, data objects can be misclassified and allocated to intuitively less desired cluster when simple matching dissimilarity measure is adopted.

2. *Center Selection*: In solving Problem $P_2$ for *k*-modes clustering, cluster centers are selected according to Theorem 2, i.e., cluster center is the frequency mode of attribute values in each cluster. However, the minimum solution $\hat{Z}$ is not unique, so $z_{i,l}$ may arbitrarily set to be the first mode of attribute values. This problem occurs frequently when clusters have weak intra-similarities, i.e., the attribute modes do not have high frequencies. In this case, it is more desirable to select those globally less frequent attribute modes as cluster centers from the viewpoint of statistical significance. That is, attribute modes with smaller frequency counts in the whole dataset should be considered with higher priority. For instance, in the Example of Section 1, attribute values *r*, *s* and *t* of *A3* in cluster 1 are candidates for mode since they have equal frequency counts in this cluster. Clearly, priority should be given to *s* since it is less frequent than *r* and *t* (as shown in Fig.1, the frequency counts of both *r* and *t* in the whole dataset are 2, which is larger than 1, i.e., the frequency count of *s*). However, standard *k*-modes algorithm cannot deal this problem due to the use of simple matching dissimilarity measure.

To address the above problems in *k*-modes algorithm, this section extends *k*-modes clustering algorithm by weighting attribute value matches in dissimilarity computation. More precisely, the similarity between two identical categorical values is not always one but a value-dependent weight value ranging from 0 to 1. Such modification on distance measure will help the algorithm to avoid the above-mentioned problems in *object allocation* and *center selection*. That is, the new optimization problem we are trying to minimize becomes (5), which subjects to the same conditions as (2) and (3).

$$P_w(U,Z) = \sum_{l=1}^{k}\sum_{i=1}^{n}\sum_{j=1}^{m} u_{i,l} d_w(x_{i,j}, z_{l,j}) \tag{5}$$

The distance function $d_w(x_{i,j}, z_{l,j})$ is defined as follows:

$$d_w(x_{i,j}, z_{l,j}) = \begin{cases} 1 - \omega(x_{i,j}, l) & (x_{i,j} = z_{l,j}) \\ 1 & (x_{i,j} \neq z_{l,j}) \end{cases} \tag{6}$$

where $\omega(x_{i,j}, l)$ is a weight value for $x_{i,j}$ in cluster $C_l$.

According to the definition of $d_w(\cdot)$, the distance between two distinct categorical values is still 1, while the distance between two identical categorical values becomes $1-\omega(\cdot)$. When $\omega(x_{i,j}, l)=1$ for any $1 \leq i \leq n, 1 \leq j \leq m$ and $1 \leq l \leq k$, the corresponding distance function is the same as in (4) in the original k-modes algorithm. That is, simple matching dissimilarity measure is a special case of $d_w(\cdot)$ when $\omega(x_{i,j}, l)$ is always fixed to be 1.

The optimization problem (5) in the attribute value weighting based k-modes clustering can be solved iteratively in a similar manner as in original k-modes clustering. That is, Problem $P_1$ can be solved according to Theorem 1 ($d(\cdot)$ is replaced by $d_w(\cdot)$). To solve Problem $P_2$, we have the following theorem, whose proof is similar to that of Theorem 3 in Ref. [14].

**Theorem 3**: Let $U = \widehat{U}$ be fixed, $P_w(\widehat{U}, Z)$ is minimized iff

$$z_{l,j} = a_j^{(r)}$$

where $a_j^{(r)}$ satisfies $f(a_j^{(r)} | C_l)\omega(a_j^{(r)}, l) \geq f(a_j^{(t)} | C_l)\omega(a_j^{(t)}, l), \forall t, 1 \leq t \leq p_j$.

**Proof:** Since $U = \widehat{U}$ is fixed, minimizing $\sum_{l=1}^{k}\sum_{i=1}^{n}\sum_{j=1}^{m} u_{i,l} d_w(x_{i,j}, z_{l,j})$ is equivalent to minimizing each $\sum_{i=1}^{n}\sum_{j=1}^{m} u_{i,l} d_w(x_{i,j}, z_{l,j})$ separately, for $1 \leq l \leq k$.

Moreover, since $\sum_{i=1}^{n}\sum_{j=1}^{m} u_{i,l} d_w(x_{i,j}, z_{l,j}) = \sum_{j=1}^{m}\sum_{i=1}^{n} u_{i,l} d_w(x_{i,j}, z_{l,j})$, minimizing $\sum_{i=1}^{n}\sum_{j=1}^{m} u_{i,l} d_w(x_{i,j}, z_{l,j})$ is equivalent to minimizing each $\sum_{i=1}^{n} u_{i,l} d_w(x_{i,j}, z_{l,j})$ separately, for $1 \leq j \leq m$.

When $z_{l,j} = a_j^{(t)}$, we have

$$\sum_{i=1}^{n} u_{i,l} d_w(x_{i,j}, z_{l,j}) = \sum_{i=1, x_{i,j}=a_j^{(t)}}^{n} u_{l,i}(1-\omega(a_j^{(t)}, l)) + \sum_{i=1, x_{i,j} \neq a_j^{(t)}}^{n} u_{l,i}$$

$$= f(a_j^{(t)} | C_l)(1-\omega(a_j^{(t)}, l)) + (|C_l| - f(a_j^{(t)} | C_l))$$

$$= |C_l| - f(a_j^{(t)} | C_l)\omega(a_j^{(t)}, l)$$

Since the size of cluster $|C_l|$ is fixed, $\sum_{i=1}^{n} u_{i,l} d_w(x_{i,j}, z_{l,j})$ is minimized iff $f(a_j^{(t)} | C_l) \omega(a_j^{(t)}, l)$ is maximal. The result follows. □

By comparing the results in Theorems 2 and 3, the cluster centers Z are updated in a different manner since we use different distance functions in (4) and (6) respectively. According to Theorem 3, the component of each cluster mode is determined by both frequency count and weight of corresponding attribute value. Therefore, *center selection* procedure could be enhanced by the use of attribute value weighting based dissimilarity measure. We remark that *object allocation* problem could also be alleviated since objects are assigned according to the new distance measure in Problem $P_1$. More specific and detailed discussion and comments will be given in Section 4.2.

## 4.2 Weighting Schemas

This section presents several frequency-based weighting methods, which result in several variants of attribute value weighting based *k*-modes algorithm. New *k*-modes algorithms with frequency-based weighting schemas are easy to be implemented since *only* minor modifications are needed on the original *k*-modes algorithm. More importantly, these new algorithms can achieve better clustering accuracies without sacrificing the scalability of original *k*-modes algorithm.

### 4.2.1 Relative Value Frequency Based Weighting Schema

The main idea is to use relative frequencies of attribute values in each cluster as the weights, i.e., the weighting function is defined as:

$$\omega(a_j^{(r)}, l) = f(a_j^{(t)} | C_l) / |C_l| \tag{WF$_1$}$$

The effect of this weighting function on cluster analysis could be investigated from two perspectives.

In allocating data objects to clusters in Problem $P_1$, the distance between two matched attribute values becomes $1 - f(a_j^{(t)} | C_l) / |C_l|$, which is a more meaningful distance function since the weak intra-similarity is taken into account. Hence, incorrect object allocation problem could be alleviated under this weighting schema. Continuing the Example of Section 1, the distance between *Y* and *mode1* and *mode2* are computed as $(1 - 3/3) + (1 - 3/3) + 1 = 1$ and $(1 - 2/3) + (1 - 2/3) + 1 = 5/3$, respectively. Therefore, cluster 1 is the nearest cluster to data object *Y*. That is, with the use of this new weighting function, the data object *Y* will be assigned to the proper cluster.

Cluster centers are selected according to Theorem 3 in Problem $P_2$, the selection criteria becomes $f^2(a_j^{(t)} | C_l) / |C_l|$. Since $|C_l|$ is fixed, $f^2(a_j^{(t)} | C_l) / |C_l|$ reaches its

maximal value iff $f(a_j^{(t)} | C_l)$ is maximal. It implies that the cluster center still is the frequency mode of attribute values in each cluster, which is the same as in the original *k*-modes algorithm. In other words, the problem in center selection still exists even the weighting function (WF$_1$) is used.

Overall, such relative value frequency based weighting schema can provide an improved procedure for object allocation, while it lacks of the capability for better center selection. In contrast, another weighting schema introduced in Section 4.2.2 is good at center selection but cannot provide much help on effective object allocation.

Another remark on weighting function (WF$_1$) is that it is *dynamic*, i.e., the weight of each attribute value is not fixed but changed dynamically in the clustering process. It is not difficult to verify such property since both $f(a_j^{(t)} | C_l)$ and $|C_l|$ change their values frequently in different iterations. Furthermore, it should be noted the same attribute value $a_j^{(t)}$ might take different weight values in different clusters.

Clearly, the use of proposed weighting schema introduced a new variant of *k*-modes algorithm, which is denoted by *df-k-modes*. The *df-k-modes* algorithm can use the same procedure as in original *k*-modes. The only difference is that we need to count and store $f(a_j^{(t)} | C_l)$ and $|C_l|$ in each iteration for the distance function evaluation. Therefore, the scalability of original *k*-modes algorithm is preserved.

**4.2.2 Uncommon Attribute Value Matches Based Weighting Schema**

This section exploits another weighting schema that is introduced by Goodall [19] for weighting uncommon attribute value matches. The Goodall measure was first proposed for biological and genetic taxonomy problems, where unusual characteristics shared by biological entities is often attributed to closely related genetic information resulting in these entities being classified into the same species [19]. Li and Biswas [6] have extended it to clustering problems in more general domains. Therefore, we can adopt the method given by Li and Biswas [6].

A pair of objects ($X_i$, $X_j$) is considered more similar than a second pair of objects ($X_p$, $X_q$), if and only if the objects $X_i$ and $X_j$ exhibit a greater match in attribute values that are less common in the population. In other words, similarity among objects is decided by the un-commonality of their attribute value matches. Similarity computed using the heuristic of weighting uncommon attribute value matches helps to define more cohesive, tight clusters where objects grouped into the same cluster are likely to share special and characteristic attribute values. One should note that common attributes values also play an important role in the similarity computation and in the clustering process. The similarity computation is realized by weighting attribute value matches between a pair of objects by the frequency of occurrence of the attribute value in the dataset.

For each attribute value $a_j^{(r)}$ in $Dom(A_j) = \{a_j^{(1)}, a_j^{(2)}, ..., a_j^{(p_j)}\}$, the *More Similar Attribute Value Set* of $a_j^{(r)}$ is defined as:

$$MSAVS(a_j^{(r)}) = \{a_j^{(t)} \mid f(a_j^{(t)} \mid D) \leq f(a_j^{(r)} \mid D), a_j^{(t)} \in Dom(A_j)\}$$

where $f(a_j^{(r)} \mid D)$ is the frequency count of attribute value $a_j^{(r)}$ in dataset $D$, i.e.,

$$f(a_j^{(r)} \mid D) = |\{x_{i,j} \mid x_{i,j} = a_j^{(r)}\}|$$

This is the set of attribute values with lower or equal frequencies of occurrence than that of $a_j^{(r)}$. Note that a value pair is more similar if it has lower frequency of occurrence. The weighting function is defined as:

$$\omega(a_j^{(r)}, l) = 1 - \sum_{a_j^{(t)} \in MSAVS(a_j^{(r)})} \frac{f(a_j^{(t)} \mid D)(f(a_j^{(t)} \mid D) - 1)}{n(n-1)} \qquad \textbf{(WF}_2\textbf{)}$$

where $n$ is the total number of objects in the dataset. If the above weighting function is used, the uncommon value matches in similarity computation will make more contribution to similarity values.

Taking the Example in Section 1 to illustrate the computation. Considering attribute *A3* in the given dataset, $f(r \mid D) = 2$, $f(t \mid D) = 2$, $f(s \mid D) = 1$ and $f(k \mid D) = 1$, i.e., values *s* and *k* are less frequent than values *r* and *t*. The *MSAVS* for attribute value *r* and *t* is:

*MSAVS* (*r*) = *MSAVS* (*t*) = {*r, t, s, k*}.

Given the *MSAVS*, the *weights* of attribute value *r* and *t* are calculated as:

$$\omega(r, l) = \omega(t, l) = 1 - (\frac{2(2-1)}{6(6-1)} + \frac{2(2-1)}{6(6-1)} + \frac{1(1-1)}{6(6-1)} + \frac{1(1-1)}{6(6-1)}) = 1 - 0.133 = 0.867.$$

Similarly, the *MSAVS* for attribute value *s and k* is:

*MSAVS* (*s*) = *MSAVS* (*k*) = {*s, k*}.

Hence, $\omega(s, l) = \omega(k, l) = 1 - (\frac{1(1-1)}{6(6-1)} + \frac{1(1-1)}{6(6-1)}) = 1$.

Therefore, attribute *A3* contributes a value of 0.867 to the similarity between objects when attribute value *r* or *t* is matched. Similarly, attribute *A3* contributes a value of 1 to the similarity between objects if attribute value *s* or *k* is matched.

Compared to the weighting function defined in Section 4.2.1, the weighting function introduced in this Section is *static*, i.e., the weight of each attribute value is fixed throughout the clustering process. That is, the weight of each attribute value can be considered to be a constant. The effect of this weighting function on cluster analysis could also be studied from two viewpoints.

In solving Problem $P_1$, the distance between two matched attribute values becomes $\sum_{a_j^{(t)} \in MSAVS(a_j^{(r)})} \frac{f(a_j^{(t)} \mid D)(f(a_j^{(t)} \mid D) - 1)}{n(n-1)}$, which is a fixed constant for each $a_j^{(r)}$. Hence, the known problem in object allocation still exists under this weighting schema.

In solving Problem $P_2$, the criteria for center selection becomes

$f(a_j^{(r)}|C_l)(1-\sum_{a_j^{(t)}\in MSAVS(a_j^{(r)})}\frac{f(a_j^{(t)}|D)(f(a_j^{(t)}|D)-1)}{n(n-1)})$. Hence, when there are multiple candidate attribute values for attribute mode, candidates with smaller global frequency counts will be selected with higher priority according to the new center selection criteria. For instance, in the Example of Section 1, attribute values *r*, *s* and *t* of *A3* in cluster 1 are valid candidates for attribute mode in the original *k*-modes algorithm since they have equal frequency in this cluster. With the use of new weighting function (WF$_2$), attribute value *s* will be selected as the representative for *A3* in cluster 1 since it makes the criteria function to reach maximal value, i.e., $(1/3)\times 1 = 1/3$. Hence, the new weighting function provides an improved center selection procedure, which avoids the problem in center selection in original *k*-modes algorithm.

From the above analysis, we can see that the uncommon value matches based weighting schema is only effective on meaningful center selection, while the incorrect data allocation problem remains unsolved.

The new variant of *k*-modes algorithm under uncommon value matches based weighting schema is denoted by *sf-k*-modes. The *sf-k*-modes algorithm can use the same procedure as in original *k*-modes algorithm after pre-computing the weights of attribute values. We note that this preprocessing step can be finished in $O(n)$ since one single pass over the data is sufficient to get all frequency counts of attribute values. Therefore, the *sf-k*-modes algorithm deserves good scalability.

### 4.2.3 Two Hybrid Weighting Schemas

As shown in previous sections, both the relative value frequency based weighting schema and uncommon value matches based weighting schema deserve certain drawbacks. Fortunately, they are complementary to each other in object allocation and center selection. It is a very natural idea to develop a new weighting function by combining these two methods. The new hybrid weighting function is defined as:

$$\omega(a_j^{(r)},l)=\frac{f(a_j^{(r)}|C_l)}{|C_l|}(1-\sum_{a_j^{(t)}\in MSAVS(a_j^{(r)})}\frac{f(a_j^{(t)}|D)(f(a_j^{(t)}|D)-1)}{n(n-1)}) \quad \textbf{(WF}_3\textbf{)}$$

The effect of above weighting function on object allocation and center selection can be analyzed in a similar manner as in previous sections. It is easy to see that this new weighting function is able to provide better solution for both object allocation and center selection, which are not well addressed in the original *k*-modes algorithm.

Motivated by the success of (WF$_3$), other kinds of hybrid weighting functions are also possible. For instance, another more efficient yet simple hybrid weighting function (WF$_4$) is defined as follows.

$$\omega(a_j^{(r)},l)=\frac{f(a_j^{(r)}|C_l)}{|C_l|f(a_j^{(r)}|D)} \quad \textbf{(WF}_4\textbf{)}$$

(WF$_4$) can be considered as a simplified version of (WF$_3$) by replacing

$$(1- \sum_{a_j^{(t)} \in MSAVS(a_j^{(r)})} \frac{f(a_j^{(t)} | D)(f(a_j^{(t)} | D)-1)}{n(n-1)}) \quad \text{with} \quad \frac{1}{|f(a_j^{(r)} | D)|}.$$

The use of (WF$_3$) and (WF$_4$) give two new variants of *k*-modes algorithm, which are denoted by *hcf-k-modes* and *hsf-k-modes*, respectively.

## 4.3 Computational Complexity

In *sf-k-modes*, *hcf-k-modes* and *hsf-k-modes*, one preprocessing step is required before entering the standard iteration process, which can be finished in $O(n)$ since one single pass over the data is sufficient to get all frequency counts of attribute values.

Hence, the time complexities of all these attribute value weighting based *k*-modes algorithms are $O(tmnk)$, where *t* is the total number of iterations required, *k* is the number of clusters, *m* is the number of attributes, and *n* is the number of objects.

The above analysis shows that these new algorithms are suitable for clustering large categorical data.

# 5. Experimental Results

A comprehensive performance study has been conducted to evaluate our methods. In this section, we describe those experiments and the results. We ran attribute value weighting based *k*-modes algorithms on real-life datasets obtained from the UCI Machine Learning Repository [20] to test their clustering performance against original *k*-modes algorithm.

## 5.1 Real Life Datasets and Evaluation Method

Six data sets from the UCI Repository are used, all of which contains only categorical attributes and class attributes. The information about the data sets is tabulated in Table 1. Note that the class attributes of the data have not been used in the clustering process.

**Table 1**. Datasets used in experiments

| Data set | Size | Attribute | Class | Class Distribution |
|---|---|---|---|---|
| Voting | 435 | 16 | 2 | 168/267 |
| Breast cancer | 699 | 9 | 2 | 241/458 |
| Mushroom | 8124 | 22 | 2 | 3916/4208 |
| Soybean | 47 | 35 | 4 | 10/10/10/17 |
| Lymphography | 148 | 18 | 4 | 2/4/61/81 |
| Zoo | 101 | 17 | 7 | 4/5/8/10/13/20/41 |

Validating clustering results is a non-trivial task. In the presence of true labels, as in the case of the data sets we used, the clustering accuracy for measuring the clustering results was computed as follows [1]. Given the final number of clusters, *k*, clustering accuracy *r* was defined as: *r*

$$=\frac{\sum_{l=1}^{k} s_l}{n}$$, where $n$ is the number of objects in the dataset, $s_l$ is the number of instances occurring in both cluster $C_l$ and its corresponding class, which had the maximal value. In other words, $s_l$ is the number of objects with the class label that dominates cluster $C_l$.

The intuition behind clustering accuracy defined above is that clusterings with "pure" clusters, i.e., clusters in which all objects have the same class label, are preferable. That is, if a partition has clustering accuracy equal to 100%, it means that it contains only pure clusters. These kinds of clusters are also interesting from a practical perspective. Hence, we can conclude that larger clustering accuracy implies better clustering results in real world applications.

## 5.2 Results

For each dataset, the number of clusters is set to be the known number of its class labels. For instance, the number of clusters is set to be 2 on voting data. We carried out 100 random runs of the original *k*-modes and new *k*-modes algorithms on each data set. In each run, the same initial cluster centers were used in all algorithms. The average clustering accuracies of different algorithms were compared.

Table 2 lists the average accuracy of clustering achieved by each algorithm over 100 runs for the six data sets. From Table 2, some important observations are summarized as follows.

(1) Firstly, it is evident from Table 2 that all attribute value weighting based *k*-modes algorithms give better clustering accuracy in comparison to standard *k*-modes algorithm. Hence, we can conclude that clustering accuracy could be greatly improved with the use of attribute value weighting technique in *k*-modes clustering.

(2) Secondly, the clustering accuracies achieved by two algorithms with basic weighting schemas (*df-k*-modes and *sf-k*-modes) are always better than that of standard *k*-modes on every dataset. It implies that original *k*-modes algorithm can be enhanced even only object allocation or center selection is given special attention.

(3) Finally, just as we have expected, two algorithms with hybrid weighting schemas (*hcf-k*-modes and *hsf-k*-modes) are more accurate than the basic weighting schema based algorithms.

**Table 2**. Average clustering accuracy (%) achieved by five algorithms on six datasets

| Data set | Standard *k*-modes | *df-k*-modes | *sf-k*-modes | *hcf-k*-modes | *hsf-k*-modes |
|---|---|---|---|---|---|
| Voting | 85.92 | 86.58 | **87.34** | 87.01 | 87.02 |
| Breast cancer | 85.00 | 86.43 | 86.87 | 91.19 | **94.91** |
| Mushroom | 73.81 | 74.61 | **76.44** | 72.70 | 72.76 |
| Soybean | 81.94 | 89.34 | 86.00 | 93.09 | **94.11** |
| Lymphography | 66.08 | 69.99 | 68.50 | **71.92** | 71.91 |
| Zoo | 82.92 | **87.30** | 84.21 | 86.23 | 84.63 |
| *Avg.* | *79.28* | *82.38* | *81.56* | *83.69* | *84.22* |

Moreover, to test the scalability of the proposed algorithms on large data sets, we applied the five algorithms to the Nursery data set [20]. This dataset consists of 12,960 data objects, where each data object is composed of 9 categorical attributes.

We tested two types of scalability. The first one is the scalability against the number of objects for a given number of clusters and the second is the scalability against the number of clusters for a given number of objects. All algorithms were implemented in Java. All experiments were conducted on a Pentium4-3.0G machine with 1G of RAM and running Windows XP. Fig. 2 shows the results of using different algorithms to find 10 clusters with different number of objects. Fig. 3 shows the results of all algorithms on finding different number of clusters from Nursery dataset.

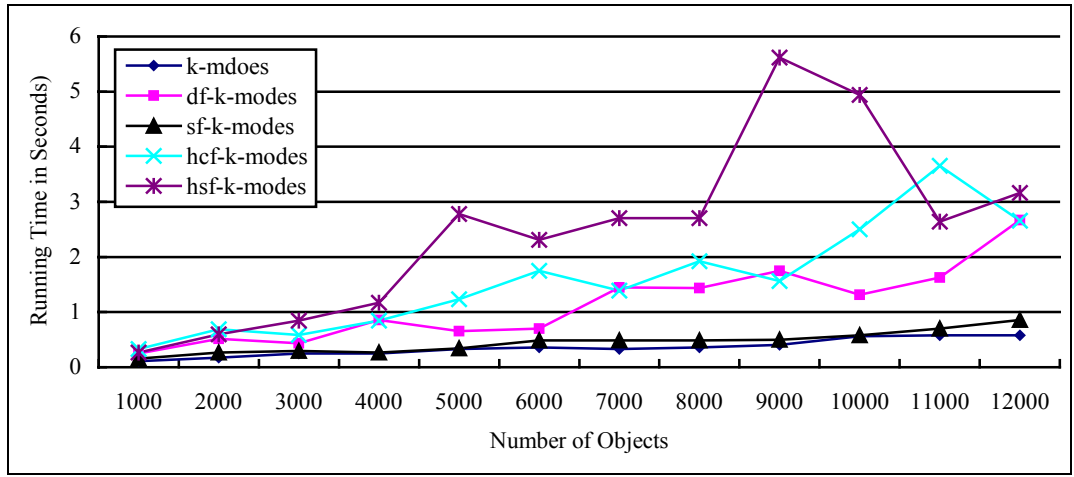

**Fig.2** Scalability to the number of objects when clustering Nursery dataset into 10 clusters

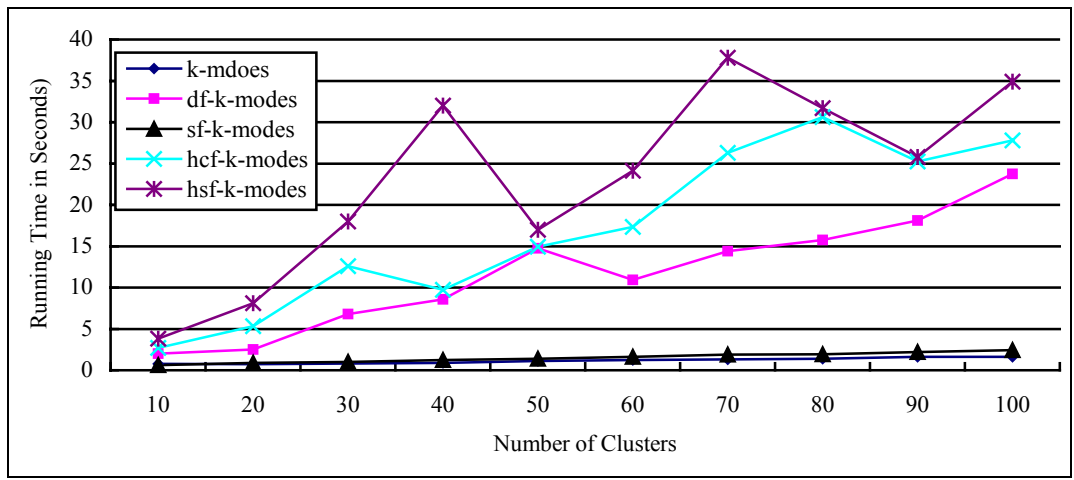

**Fig.3** Scalability to the number of clusters when clustering 12,960 objects of Nursery dataset

One important observation from these figures was that all algorithms show a linear increase in the running time as the number of data objects and the number of clusters is increased, which is highly desired in real world applications.

Furthermore, although all variants of attribute value weighting based *k*-modes algorithms are slower than the original *k*-modes algorithm, it should be noted that their performance are comparable. In particular, the running time of *sf-k*-modes algorithm is almost the same with that of the *k*-modes algorithm.

We also see that *hcf-k*-modes, *hsf-k*-modes and *df-k*-modes are relatively slower than *sf-k*-modes. The reason behind is that *hcf-k*-modes, *hsf-k*-modes and *df-k*-modes take much more iterations to find better solutions.

In summary, the above experiments show that our new algorithms achieve better clustering accuracies without sacrificing the scalability of original *k*-modes algorithm too much.

# 6. Conclusions

The conventional *k*-modes algorithm is efficient and effective in clustering large categorical data. However, its use of a simple matching dissimilarity measure compromises its effectiveness and its ability to correctly classify categorical data. Therefore, this paper extends *k*-modes clustering algorithm by weighting attribute value matches in dissimilarity computation. New variants of *k*-modes algorithm are easy to be implemented with only minor modifications on the original *k*-modes algorithm. More importantly, these new algorithms can achieve better clustering accuracies without sacrificing the scalability of *k*-modes algorithm. The superiority of these algorithms against standard *k*-modes algorithm was demonstrated through several experiments.

## Acknowledgements

This work was supported by the High Technology Research and Development Program of China ((No. 2004AA413010, No. 2004AA413030), the National Nature Science Foundation of China (No. 40301038) and the IBM SUR Research Fund.

## References

[1] Z. Huang. Extensions to the k-Means Algorithm for Clustering Large Data Sets with Categorical Values. Data Mining and Knowledge Discovery, 1998, 2: 283-304

[2] C. R. Palmer, C. Faloutsos. Electricity Based External Similarity of Categorical Attributes. In: Proc. of the 7th Pacific-Asia Conference on Advances in Knowledge Discovery and Data Mining (PAKDD'03), pp.486-500, 2003

[3] S. Q. Le, T. B. Ho. A Conditional Probability Distribution-Based Dissimilarity Measure for Categorical Data. In: Proc. of the 8th Pacific-Asia Conference on Advances in Knowledge Discovery and Data Mining (PAKDD'04), pp. 580-589, 2004

[4] V. Cheng, C. H. Li, J. T. Kwok, C-K. Li. Dissimilarity learning for nominal data. Pattern Recognition, 2004, 37(7): 1471-1477

[5] S-G. Lee, D-K. Yun. Clustering Categorical and Numerical Data: A New Procedure Using Multidimensional Scaling. International Journal of Information Technology and Decision Making, 2003, 2(1): 135-160

[6] C. Li, G. Biswas. Unsupervised learning with mixed numeric and nominal data. IEEE Transactions on Knowledge and Data Engineering, 2002, 14(4): 673-690.

[7] Z. Huang, M. K. Ng. A fuzzy k-modes algorithm for clustering categorical data. IEEE Transaction

on Fuzzy Systems, 1999, 7(4): 446-452.

[8]  M. K. Ng, J. C. Wong. Clustering categorical data sets using tabu search techniques. Pattern Recognition, 2002, 35(12): 2783-2790.

[9]  G. Gan, Z. Yang, J. Wu. A Genetic *k*-Modes Algorithm for Clustering Categorical Data. In: Proc. of ADMA'05, pp.195-202, 2005

[10] D-W. Kim, K. Y. Lee, D. Lee, K. H. Lee. A *k*-populations algorithm for clustering categorical data. Pattern Recognition, 2005, 38(7): 1131-1134.

[11] Dae-Won Kim, K. H. Lee, D. Lee. Fuzzy clustering of categorical data using fuzzy centroids. Pattern Recognition Letters, 2004, 25(11): 1263-1271.

[12] Y. Sun, Q. Zhu, Z. Chen. An iterative initial-points refinement algorithm for categorical data clustering. Pattern Recognition Letters, 2002, 23(7): 875-884.

[13] Z. He, S. Deng, X. Xu. Improving k-modes algorithm considering frequencies of attribute values in mode. In: Proc. of International Conference on Computational Intelligence and Security, LNAI 3801, pp. 157-162, 2005.

[14] M. K. Ng, M .J. Li, J. Z. Huang, Z.He. On the Impact of Dissimilarity Measure in k-Modes Clustering Algorithm. IEEE Transactions on Pattern Analysis and Machine Intelligence, 2007.

[15] D.S. Modha, W.S. Spangler. Feature Weighting in k-Means Clustering. Machine Learning, 2003, 52(3): 217-237

[16] H. Frigui, O. Nasraoui. Unsupervised learning of prototypes and attribute weights. Pattern Recognition, 2004, 37(3): 567-581

[17] E. Y. Chan, W-K. Ching, M. K. Ng, J. Z. Huang. An optimization algorithm for clustering using weighted dissimilarity measures. Pattern Recognition, 204, 37(5): 943-952.

[18] J. Z. Huang, M. K. Ng, H. Rong, Z. Li. Automated Variable Weighting in k-Means Type Clustering. IEEE Transactions on Pattern Analysis and Machine Intelligence, 2005, 27(5): 657-668

[19] D. W. Goodall. A New Similarity Index Based on Probability. Biometrics, 1966, 22: 882-907.

[20] Merz, C. J., Merphy. P.: UCI Repository of Machine Learning Databases. [http://www.ics.uci.edu/~mlearn/MLRRepository.html] (1996)